# Underwater Fish Species Classification using Convolutional Neural Network and Deep Learning

Dhruv Rathi*, Sushant Jain**, Dr. S. Indu***

* Department of Computer Engineering, Delhi Technological University, New Delhi, India
Phone: +919868904222, E-Mail: dhruvrathi15@gmail.com, http://dhruvrathi.me/
** Department of Computer Engineering, Delhi Technological University, New Delhi, India
E-Mail: sjsushantjain22@gmail.com
*** Department of Electronics and Communication, Delhi Technological University, New Delhi, India E-Mail: s.indu@dce.ac.in, http://www.dtu.ac.in/Web/Departments/Electronics/faculty/sindu.php

*Abstract*— **The target of this paper is to recommend a way for Automated classification of Fish species. A high accuracy fish classification is required for greater understanding of fish behavior in Ichthyology and by marine biologists. Maintaining a ledger of the number of fishes per species and marking the endangered species in large and small water bodies is required by concerned institutions. Majority of available methods focus on classification of fishes outside of water because underwater classification poses challenges such as background noises, distortion of images, the presence of other water bodies in images, image quality and occlusion. This method uses a novel technique based on Convolutional Neural Networks, Deep Learning and Image Processing to achieve an accuracy of 96.29%. This method ensures considerably discrimination accuracy improvements than the previously proposed methods.**

## I. INTRODUCTION

Monitoring the behavior of different species of fishes is of primary importance for getting insights into a marine ecological system. The count and distribution of the various species of fishes can give valuable insights about the health of the ecological system and can be used as a parameter for monitoring environmental changes. Visual classifying of fishes can also help trace their movement and give patterns and trends in their activities providing a deeper [1]knowledge about the species as a whole.

The study of the behavior of the fishes can be automated by getting visual feedback from multiple locations and automating the process of visual classification of fishes, which will give significantly larger amounts of data for pattern recognition. Although, there have been many advances in classifying fish taken out of water [1][2][3][4] or in artificial conditions, such as in tanks with adequate lighting [5], there has been no significant breakthrough in the classification of fishes in datasets created from underwater videos. The challenges faced during underwater classification of fish species include noise, distortion, overlap, segmentation error and occlusion [6]. Also, the complex environment restricts simpler approaches like luminance, and background subtraction includes issues such as colors shifting, inconsistent lighting and presence of sediments in water and undulating underwater plants.

Fish species recognition is a multi-class classification problem and is a compelling research field of machine learning and computer vision. The state-of-the-art algorithms implemented over individual input images which perform classification mainly using shape and texture feature extraction and matching [7][8]. All the existing work either deals with a small dataset distinguishing between less number of species or has a low accuracy. Our proposed method uses Convolutional Neural Networks which makes the process simpler and more robust even while working with a large dataset. CNNs are also much more flexible and can adapt to the new incoming data as the dataset matures. We make use of the fish dataset from the Fish4Knowledge project [24] for testing our algorithm. We perform the classification by pre-processing the images using Gaussian Blurring, Morphological Operations, Otsu's Thresholding and Pyramid Mean Shifting, further feeding the enhanced images to a Convolutional Neural Network for classification.

The remaining paper is catalogued as follows: Section II. reviews the current research done in this field; Section III. delineates the proposed algorithm; Section IV. gives the experimental results; Section V. discusses the Future Works and Conclusion and Section VI. gives the References.

## II. RELATED WORKS

There are numerous approaches proposed by researchers for the classification of fish species as delineated below:

C. Spampianto et al. [7] attempt to classify fishes using texture features extracted from the gabor filtering, gray level histogram and the histogram of Fourier descriptors of boundaries and Curvature Scale Space Transform used for shape features extracted using. The algorithm was tested on a dataset of 360 images and achieved an accuracy of 92%.

Takakazu Ishimatsu et al. [8] used two identification features: speckle patterns and scale forms of fish and used morphological algorithms and filters for discrimination. They just showed the differentiation between three species-Pilchard Sardine (ma-iwashi), Japanese Horse Mackerel (ma-aji), and Common Mackerel (ma-saba) with an accuracy of 90%, 88% and 90% respectively. Their method is dependent on the size and shape of the morphological filters used.

Junguk Cho et al. [9] used Haar classifiers with the Scythe Butterfly fish as their test species. Their method is heavily dependent on the background environment in the image of the fish and the angle from which the photograph is taken.

Andrew Rova et al. [10] attempt to classify fishes using the method of warping the images prior to classification using SVM on a dataset of 320 images achieving an accuracy of 90%.

Chomptip Pornpanomchai et al. [11] proposed and developed a fish recognition system based on shape and texture. They compared Artificial Neural Networks and Euclidean Distance Method (EDM) working on a test set of 300 images and a training set of 600 images. They achieved an accuracy of 81.67% and 99.00% with EDM and ANN respectively.

Rodrigues et al. [13] recommended an algorithm based on SIFT feature extraction & Principal Component Analysis (PCA) but they worked on a very small dataset of 162 images encompassing 6 different species getting an accuracy of 92%.

S. Sclaroff et al. [14] performed object deformable shape detection and object detection was done through model-based region grouping. Computational complexity is the major drawback of this method.

C. Spampinato et al. [15] worked on 20 underwater videos to detect, track and count fishes with an accuracy of 85%. They performed detection using Moving Average Detection Algorithm and Adaptive Gaussian Mixture Model.

Andres Hernandez-Serna et al. [16] used Artificial Neural Networks for the automatic identification of species. They worked on a dataset of 697 images achieving an accuracy of 91.65%.

Arjun Kumar Joginpelly et al. [17] propose an automatic technique using Gabor filters to extract important features from two species, Epinephelus morio and Ocyurus chrysurus. The proposed algorithm is tested on 200 frames, each containing many fish and non-fish regions. The accuracy is 70.6% for Epinephelus morio and 80.3% for Ocyurus chrysurus.

Deokjin Joo et al. [18] extracted stripe and color patterns of wild cichlids and used Random Forests and SVM for classification achieving an accuracy of 72% on a dataset of 594 wild cichlids. They have a low accuracy and they just target Cichlid fishes.

Yi-Haur Shiau et al. [19] proposed a method of sparse representation-based classification for the recognition and verification of fishes which maximizes the probability of partial rankings thus obtained. They worked on a dataset of 1000 images and achieved the highest accuracy of 81.8% for a particular feature space dimensionality.

S.O. Ogunlana et al. [20] classified using Support Vector Machine technique based on the shape features of the fish. They worked on a training data of 76 fish and testing data of 74 fish achieving an accuracy of 78.59%.

S. Cadieux et al. [21], generates the contours of fish in an unconstrained environment by deploying an infrared silhouette sensor which acquires contours of fish in a constrained flow. When the inputs are noisy, these features give a poor performance. The system has reported classification accuracy around 78%.

D. J. Lee et al. [22], removed edge noise and redundant data points through the development of a shape analysis algorithm. Critical landmark points were located using an algorithm of curvature function analysis. A group of nine fish species was used to test this method. The dataset that was used to perform the experimentation consisted of only 22 images.

M. Nery et al. [23], proposed a methodology based on feature selection. This method develops a feature vector by utilizing a set of descriptors obtained by analysis of the characteristic contribution of an individual descriptor to the overall performance for classification. A classification accuracy of about 85% is reported.

### III. PROPOSED METHODOLOGY

Here, we present a methodology for the discrimination of fish species. The dataset used for the concerned work is taken from [24]. The initial step taken by the system aims at removing the noise in the dataset. Application of Image Processing before the training step helps to remove the underwater obstacles, dirt and non-fish bodies in the images. The second step uses Deep Learning approach by implementation of Convolutional Neural Networks(CNN) for the classification of the Fish Species.

In order to get the best results for feature identification and training of the CNN, it is important to provide input image with enhanced features as training sample. The pre-processing consists of the following steps,

```
kernel = np.ones((3,3),np.uint8)
opening = cv2.morphologyEx(thresh,cv2.MORPH_OPEN,kernel, iterations = 2)
sure_bg = cv2.dilate(opening,kernel,iterations=3)
dist_transform = cv2.distanceTransform(opening,0,5)

ret, sure_fg = cv2.threshold(dist_transform,0.7*dist_transform.max(),255,0)
sure_fg = np.uint8(sure_fg)
unknown = cv2.subtract(sure_bg,sure_fg)
```

With the implementation of Otsu's thresholding, a grey level histogram is created from the grayscale image for noise removal, if a pixel of the grayscale is greater than the threshold value, it is considered to be white, else declared as black. The image provides a sure foreground with the fish in focus.

Next step deals with the implementation of Morphological Operations, viz-a-viz, Erosion and Dilation of the binarized image.

In the Erosion of the image, a kernel, precisely a fixed size matrix is convolved over the image. A pixel in the processed image will be taken as 1 only if all the pixels under the kernel are 1, otherwise, the pixel is eroded (0).

Thus, in this step, the thickness of the foreground object(fish) decreases.

The step of Dilation implements the algorithm that a pixel in the processed image is 1 if at least one pixel under the kernel is 1.

Thus, this step is used to join the broken parts of the image from the noise removal step.

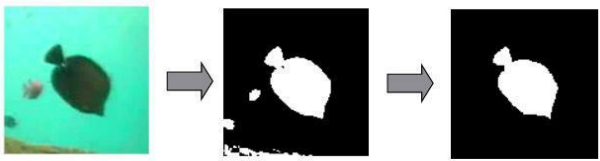

Figure 1. Pre-processing with Otsu's binarization, Dilation and Erosion.

The Second step of the procedure is the implementation of a Convolutional Neural Network(CNN) for classification of Fish species.

The input layer of the network takes the 100x100x3 original RGB image stacked with the 100x100x1 image which is the output of the pre-processing stage, thus making the input of 100x100x4, the fully-connected layer where we get the trained output and the intermediate hidden layers. The network has a series of convolutional and pooling layers. Neurons in layer say, 'm' are connected to a subset of neurons from the previous layer of (m-1), where the (m-1) layered neurons have contiguous receptive fields, as shown in Fig(2a).

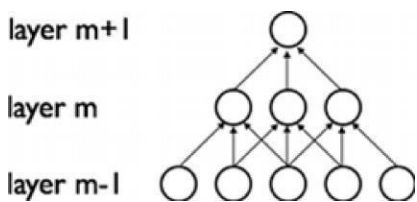

Figure 2a. Graphical flow of layers showing the connection between layers.

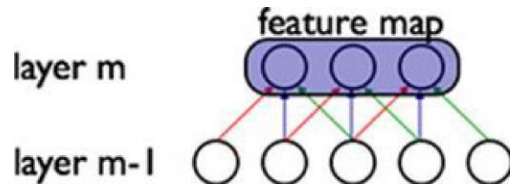

Figure 2b. Graphical flow of layers showing sharing of Weights.

Fig (2b) represents three hidden units. The weights of similar color are shared, thus are inferred to be identical.

The summation of the gradients of the parameters that are being shared results in the gradient of the shared weight. Such similarity thus allows detection of features regardless of their positions in the visual field. In addition to this, weight sharing tends to decrease the number of free learning parameters. Due to this control, CNN tends to achieve better generalization on vision problems.

The Max-pooling layers act as non-linear down sampling, in which the input image is partitioned into non-overlapping rectangles. The output of each sub-region is the maximum value.

The Convolution Layer is the first layer of the CNN network. The structure of this layer is shown in fig (3). It consists of a convolution mask, bias terms and a function expression. Together, these generate the output of the layer. The figure below shows a 5x5x4 mask that performs convolution over a 100x100x4 input feature map. Upon application of 32 such 5x5x4 filters, the resultant output is a 96x96x32 matrix.

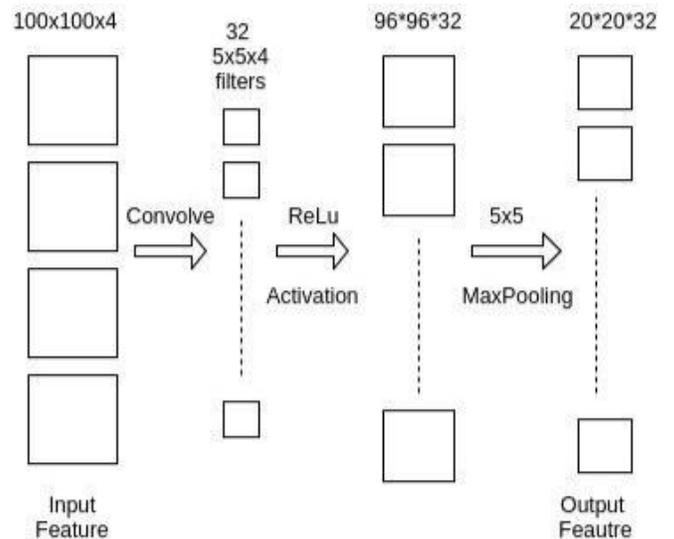

Figure 3. Processing the input feature with 32 filters and max-pooling.

The next layer in the network is a subsampling layer. The Subsampling layer is designed to have the same number of planes as the convolution layer. The purpose of this layer is to reduce the size of the feature map. It divides the image into blocks of 5x5 and performs max-pooling. Sub-sampling layer preserves the relative information between features and not the exact relation.

Figure 3 shows how the input features are processed with 32 filters and max-pooling. The above process is repeated two times again once with 64 filters and then with 32 filters.
The final output is connected to a fully connected layer which is further connected to an 80% dropout layer and lastly another fully connected layer which classifies the images into appropriate categories.

The aim of the training algorithm is to train a network such that the error is minimized between the network output and the desired output.
In the proposed method, we provide the comparison of different Activation Functions that will be applied to the different Layers in the CNN. The following are the three Activation functions, namely,

*a) ReLU  b) Sigmoid  c) tanh*

Mathematical definition of the above functions explained below:
a) **ReLU** : Rectified Linear Unit is :
$$ReLU: h = \max(0, a), \text{ where } a = W*x+b$$

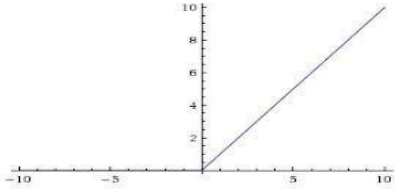

b) **Softmax**:
$$\sigma(z) = (e^z / \sum_{k=1}^{k} (e^{z_k}))$$

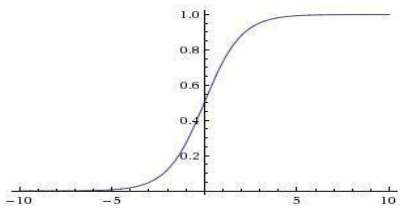

c) **tanh**: Mathematical definition states,
$$\tanh = (e^{+x} + e^{-x}) / (e^{-x} - e^{+x})$$

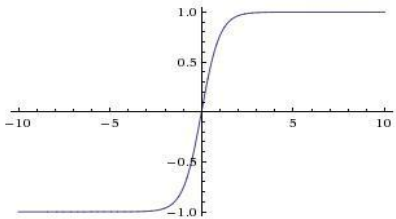

The loss function used after the fully-connected layer is Cross-entropy, Mathematically,

$$Hy'(y) := -\sum_i (y'_i \log(y_i) + (1 - y'_i) \log(1 - y_i))$$

which can be explained as a (minus) log-likelihood for the data $y'_i$, under a model $y_i$.

TABLE 1. ALGORITHM FOR PRE-PROCESSING AND CONVOLUTIONAL NETWORK:

**Algorithm 1.** Fish Classification Algorithm

**Input:**
- Training set $X = \{(X_i, c_i)\}_{i=1}^{N}$ from the given dataset undergoes Otsu's binarization and thresholding to create a grey-level histogram, further implementing Dilation, Erosion and Subtraction to remove the background noise and other water objects.
- Pyramid Mean Shifting implemented to highlight the outlines of the fish object in the image with the kernel function $k(x_i - x)$.

**Processing:**
- Architect the model for efficient mapping of features and set appropriate hyper-parameters with the input dimensions as that of X.
- Compute Feature matrices of each layer which gets more detailed with further layers and feed it as input to next layer.
- Feed the Fully Connected Layer with the final feature matrix. Repeat the previous step with the three Activation functions discussed, ReLU, tanh, Sigmoid.

**Output:**
- Using *Adam optimizer* for the Fully Connected Layer outputs

## IV. EXPERIMENTAL RESULTS

The proposed method was tested in Python on the dataset Fish4Knowledge [24] of 27,142 images shown in Figure 3 below.

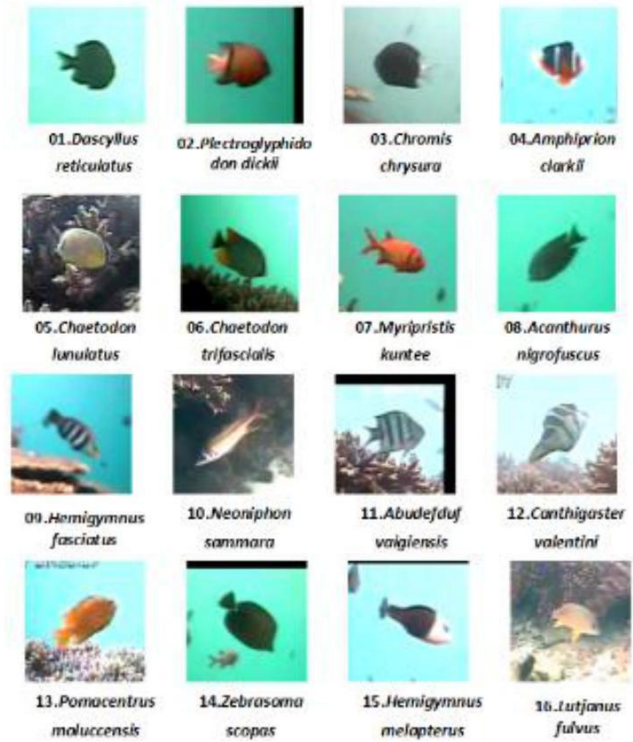

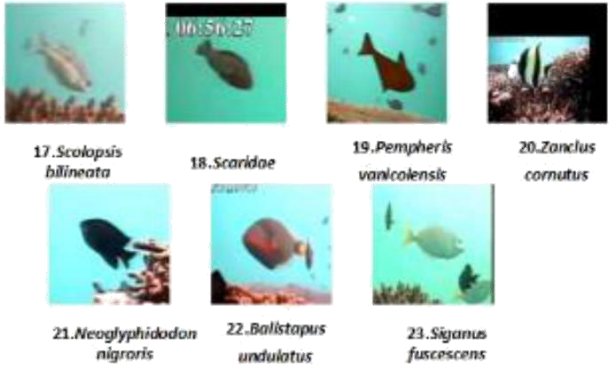
Figure 2. Dataset used

Table 2 provides insights into the number of samples used for each different species covered. Table 3 shows the results as the accuracy of the correctly predicted test images in the sample as given by Eqn. (5) shown below:

$$Accuracy = \frac{TP}{TP+FP} \times 100\% \quad - (5)$$

| Species | Number of Images |
|---|---|
| Plectroglyphidodon dickii | 11312 |
| Chromis chrysura | 2683 |
| Amphirion clarkii | 3593 |
| Chaetodon lunulatus | 4049 |
| Chaetodon trifascialls | 2534 |
| Myripristis kuntee | 190 |
| Acanthurus nigrofuscus | 450 |
| Hemigymnus fasciatus | 218 |
| Chaetodon trifascialis | 242 |
| Neoniphon sammara | 298 |
| Abudefduf vaigiensis | 198 |
| Canthigster valentini | 148 |
| Pomocentrus molucensis | 180 |
| Hemigymnus fasciatus | 190 |
| Scolopsis billineate | 142 |
| Neoniphon sammara | 206 |
| Scaridae | 149 |
| Pemphereis vanicolensis | 156 |
| Zanclus cornutus | 129 |
| Balistapus undulatus | 221 |
| Zebrasoma scopas | 116 |
| **Total** | **27,142** |

Table 2. The Species used in the Dataset

| Activation Function Used | Overall Accuracy |
|---|---|
| **ReLU** | **96.29%** |
| tanh | 72.62% |
| Softmax | 61.91% |

TABLE 3. RESULTS

## V. CONCLUSION AND FUTURE WORK

The proposed method of the classification of fish species gives an accuracy of 96.29% which is very high compared with the other current implemented methods used for this application. Hence the proposed approach can certainly be used for real time applications as the computation time is 0.00183 seconds per frame. The method couldn't achieve 100% accuracy as some images couldn't be classified accurately due to the effect of background noise and other water bodies. We plan to improvise our algorithm further by implementing Image Enhancement techniques to counter for the lost features in the images.